\documentclass[10pt]{elsarticle}

\usepackage{amssymb}
\usepackage{amsmath}
\usepackage{xurl}
\usepackage{xcolor}
\usepackage[inline]{enumitem}
\usepackage{amsthm}

\usepackage{soul}
\newcommand{\declare}{\ensuremath{\textsc{Declare}}}

\theoremstyle{definition} 
\newtheorem{definition}{Definition}[section]

\usepackage{silence}
\usepackage{titlesec}
\titleformat{\paragraph}[runin]{\normalfont\normalsize\itshape}{\theparagraph}{1em}{}
\titlespacing*{\paragraph}{0pt}{3.25ex plus 1ex minus .2ex}{1.5ex plus .2ex}

\newif\ifshowcomments
\showcommentstrue
\ifshowcomments
    \newcommand{\mynote}[2]{\fbox{\bfseries\sffamily\scriptsize{#1}}{\small$\blacktriangleright$\textsf{#2}$\blacktriangleleft$}}
\else
    \newcommand{\mynote}[2]{}
\fi

\newcommand{\rv}[1]{\textcolor{black}{{#1}}}
\newcommand{\rmv}[1]{}

\usepackage{hyperref}
\usepackage{times}
\usepackage{comment}

\journal{Information Systems}

\begin{document}

\begin{frontmatter}


\title{Agentic Business Process Management:\\ A Research Manifesto}

\author[label6]{Diego Calvanese}
\ead{diego.calvanese@unibz.it}
\affiliation[label6]{organization={Free University of Bozen-Bolzano},
            city={Bozen-Bolzano},
            country={Italy}}

\author[label8]{Angelo Casciani}
\ead{casciani@diag.uniroma1.it}

\author[label1]{Giuseppe De Giacomo}
\ead{giuseppe.degiacomo@cs.ox.ac.uk}
\affiliation[label1]{organization={University of Oxford},
            city={Oxford},
            country={United Kingdom}}

\author[label2]{Marlon Dumas}
\ead{marlon.dumas@ut.ee}
\affiliation[label2]{organization={University of Tartu},
            city={Tartu},
            country={Estonia}}

\author[label3]{Fabiana Fournier}
\ead{fabiana@il.ibm.com}
\affiliation[label3]{organization={IBM Research},
            city={Haifa},
            country={Israel}}

\author[label4,label5]{Timotheus Kampik}
\ead{tkampik@cs.umu.se}
\affiliation[label4]{organization={Umeå University},
            city={Umeå},
            country={Sweden}}

\affiliation[label5]{organization={SAP},
            city={Berlin},
            country={Germany}}

\author[label1]{Emanuele La Malfa}
\ead{emanuele.lamalfa@cs.ox.ac.uk}

\author[label3]{Lior Limonad}
\ead{liorli@il.ibm.com}

            
\author[label8]{Andrea Marrella}
\ead{marrella@diag.uniroma1.it}
\affiliation[label8]{organization={Sapienza Università di Roma},
            city={Rome},
            country={Italy}}

\author[label9]{Andreas Metzger}
\ead{andreas.metzger@paluno.uni-due.de}
\affiliation[label9]{organization={paluno (Ruhr Institute for Software Technology), University of Duisburg Essen},
            city={Essen},
            country={Germany}}

\author[label6]{Marco Montali}
\ead{marco.montali@unibz.it}

\author[label15]{Daniel Amyot}
\ead{damyot@uottawa.ca}
\affiliation[label15]{
organization={University of Ottawa},
city={Ottawa},
country={Canada}
}

\author[label13,label14]{Peter~Fettke}
\ead{peter.fettke@dfki.de}
\affiliation[label13]{
organization={German Research Center for Artificial Intelligence (DFKI)},
city={66123 Saarbrücken},
country={Germany}
}
\affiliation[label14]{
organization={Saarland University},
city={66123 Saarbrücken},
country={Germany}
}

\author[label12]{Artem~Polyvyanyy}
\ead{artem.polyvyanyy@unimelb.edu.au}
\affiliation[label12]{
organization={The University of Melbourne},
city={Parkville, VIC 3010},
country={Australia}
}

\author[label7]{Stefanie~Rinderle-Ma}
\ead{stefanie.rinderle-ma@tum.de}
\affiliation[label7]{
organization={TUM School of Computation, Information, and Technology, TU Munich},
city={Garching},
country={Germany}
}

\author[ssardina]{Sebastian Sardi\~na}
\ead{sebastian.sardina@rmit.edu.au}
\affiliation[ssardina]{
organization={RMIT University},
city={Melbourne},
country={Australia}}

\author[label11]{Niek~Tax}
\ead{niek@meta.com}
\affiliation[label11]{
organization={Meta},
city={London},
country={United Kingdom}
}

\author[label10]{Barbara Weber}
\ead{barbara.weber@unisg.ch}
\affiliation[label10]{organization={University of St. Gallen},
             city={St. Gallen},
             country={Switzerland}}

\begin{abstract}
This paper presents a manifesto that articulates the conceptual foundations of \emph{Agentic Business Process Management} (APM), an extension of Business Process Management (BPM) for governing autonomous agents executing processes in organizations.
From a management perspective, APM represents a paradigm shift from the traditional\rmv{\rv{macro-level}} view on business processes.
This shift is driven by the realization of process awareness \rv{by}\rmv{\rv{micro-level}} agent-oriented abstractions: software and human agents act as primary functional entities that perceive, reason, and act within explicit process frames. \rv{Thus, APM moves away from automation-oriented BPM toward systems} in which autonomy is constrained, aligned, and made operational through \emph{process aware} agents.

We introduce the core abstractions and architectural elements required to realize APM systems and elaborate on four key capabilities that agents in APM systems must support: \emph{framed autonomy}, \emph{explainability}, \emph{conversational actionability}, and \emph{self-modification}. These capabilities jointly ensure that agents’ goals are aligned with organizational goals and that agents behave in a framed yet proactive manner in pursuing those goals. We discuss the extent to which the capabilities can be realized and identify research challenges whose resolution requires further advances in BPM, AI, and multi-agent systems. The manifesto thus serves as a roadmap for bridging these communities and for guiding the development of APM systems in practice.
\end{abstract}




\end{frontmatter}




\sloppypar

\section{Introduction and Motivation}
\label{sec:intro}

With the advent of big data analytics,
machine learning, and foundation models---notably, Large Language Models (LLMs)---organizations increasingly adopt data-driven artificial intelligence (AI) approaches to realize software systems that form the backbone of their operations.
A long-term ambition leveraging these technologies to instill \emph{autonomy} into software systems, meaning that they independently perceive, reason, and act to achieve their goals, emphasizing intentionality, goal-directed behavior, and constrained autonomy~\cite{wooldridge1995intelligent}.
This lifts the burden of low-level decisions from knowledge workers and can ultimately reduce costs, alleviate workforce shortages, foster creativity and competitiveness, and improve the experiences of employees, customers, and other stakeholders involved~\cite{Shen2024-AIEmployment}.
Currently, efforts to achieve greater autonomy in software systems often attempt to make use of LLMs and are summarized under the umbrella term of \emph{agentic AI}, thus promoting the notion of \emph{agents} and \emph{Multi-Agent Systems} (MAS) as the fundamental abstractions of software systems~\cite{he2025llm-mas-se}.

However, increasing the autonomy of LLM-based software agents also brings risks. On the business level, such agents may make decisions that are in violation of compliance rules, disregard social norms and expectations, or simply lead to directly adverse business outcomes~\cite{Warren2025-NLWeb}. On the technical level, reliance on LLMs entails inherent functional uncertainty that is difficult to test and debug, leading to system integration and maintenance challenges and, ultimately, to technical debt when replacing software agents requires making sense of entangled behaviors emerging from agent interactions.

To mitigate and control these risks, it is crucial to \emph{govern} software agents in organizations. 
Agent governance has been a well-established line of research for several decades~\cite{10.1145/3507910}. It originated primarily in the AI subfield of \emph{normative MAS}, which focuses on regulating autonomous agent behavior through the specification of norms such as obligations, permissions, and prohibitions to ensure social order and compliance~\cite{gabbay2021handbook}.
On the more practical side, LLM providers have issued recommendations for governing LLM-based agents~\cite{Shavit2023-AgenticAI}.
Still, long-running research lines and pragmatic practice-based guidelines for governing agents lack a bridge to a practically well-established perspective of managing work in organizations.

Such a perspective is provided by Business Process Management (BPM) literature and practice. 
BPM is concerned with the management of (sets of) 
activities that are performed in coordination \rmv{(a \textit{business process}) }and jointly contribute to the accomplishment of organizational (business) goals~\cite{weske2024business,dumas2018fundamentals}.
With the advent of AI agents in modern organizations and the delegation of some activities to autonomous agents, it is imperative that BPM also encompasses the management of such agents acting within organizations.
Indeed, BPM and MAS research have a strong joint tradition, dating back to the 1990s (cf.~\cite{Vu25} for an overview).
Still, most of the corresponding work focuses on allocating agents to execute tasks within a business process, rather than on managing autonomous agents as first-class citizens to ensure the achievement of business goals.
While some exceptions exist---notably, work on process choreographies~\cite{decker2009design} and the emerging research directions of \emph{agent system mining}~\cite{DBLP:conf/bpm/TourPKS23,DBLP:journals/access/TourPK21}, as well as of \emph{agentic AI process observability}~\cite{AgenticObservability2025}, within process mining---there is a lack of a holistic perspective on what is needed to apply and extend BPM to the governance of software agents in organizations.
Moreover, BPM practitioners often lack a clear understanding of what constitutes an agent (and its associated benefits and risks), resulting in overlooked insights and missed opportunities for meaningful progress in the emerging domain of \textit{Agentic (Business) Process Management} (APM)~\cite{Vu25}.

Consider an example of an APM system that facilitates the onboarding of new suppliers as part of a procurement process~\cite{Fettke2025}.
In such a system, there is a buyer agent and several supplier agents, some of which are human and some AI-based. Both types of agents (buyer and supplier) possess process awareness in the sense that they continuously align their individual goals and act in concert to meet the organizational objectives of the procurement process. 

To address the above lack of a holistic perspective, this manifesto introduces the APM paradigm, with the notion of \emph{framed agency} at its core.
Specifically, we provide a joint position of academics focusing on both theoretical and applied aspects of the fields of BPM and autonomous agents, as well as that of industry experts.\rmv{The authors of this paper---with a strong background in BPM, process mining, software engineering, AI, and MAS---}
We identified and discussed the required capabilities of APM and associated research challenges during the Dagstuhl Seminar \#25192 (AUTOBIZ\footnote{See \url{https://www.dagstuhl.de/25192}. We express our gratitude to the Scientific Directorate and staff of Schloss Dagstuhl for their invaluable support. We also thank our fellow participants for their engaging discussions.}) and follow-up sessions. \rmv{Jointly with the seminar participants, we discussed and developed core concepts, challenges, and research directions for APM.}
Specifically, after a series of talks by experts,  participants split into working groups to further discuss individual topics of the research agenda.
The results of these breakout groups were presented to all seminar participants, and their feedback was used to improve the findings. These were further discussed during the PMAI'25 workshop\footnote{\url{https://ceur-ws.org/Vol-4087/}; intermediate proposals that this paper extends are provided in~\cite{calvanese2025autonomy,Fettke2025,Montali25,Senderovich25}.}, culminating in the vision of APM systems.

We first present the underlying concepts and core definitions of an APM system, followed by the main architectural elements, highlighting the key distinctions between the APM system level and the agent level (Section~\ref{sec:apm}).
Then, we discuss four key capabilities that are required to achieve responsible and effective management of AI agents in business processes: \emph{framed autonomy}, \emph{explainability}, \emph{conversational actionability}, and \emph{self-modification} (Section~\ref{sec:capabilities}).
These capabilities are purposefully ordered: Framing ensures that agents are process-aware and their actions are
guard-railed. Explainability serves as a means to preserve the system's autonomy by having the agents articulate the rationale for their behavior, a prerequisite
to deploying them. Conversational actionability then ensures effective execution, interaction, and governance.
Finally, self-modification allows for continuous improvement to help move
towards the long-term vision of self-improving software systems in the operational
back-ends of organizations. The realization and delivery of these four key capabilities
involves a series of research challenges, which are presented in Section~\ref{sec:challenges}.
We conclude the paper with critically reflections and call for the establishment of APM formal foundations, as well as engineering and management practices, and for empirical research to confirm or challenge the assumptions made in this manifesto (Section~\ref{sec:discussion}).

\bigskip

\section{Agentic Business Process Management Systems}
\label{sec:apm}

\noindent This section first defines the fundamental concepts underlying an APM system and then elaborates on the main architectural elements that constitute such a system at both \emph{macro} (system) and \emph{micro} (agent) levels.

\subsection{Fundamental Concepts}
\label{sec:concepts}

\noindent In line with the AI and software engineering communities~\cite{Sapkota25,Wooldridge2009,TOSEM-SI-2025}, we introduce the concept of an \textit{agentic system} as a collection of (one or more) \textit{individual} goal-driven agents that sense, reason, and act upon external stimuli to deliver (parts of) the functionality of a software system~\cite{Sapkota25,Wooldridge2009,Jennings00}.
We use the term \emph{agentic} to indicate that the APM system is agent-centric: agents constitute the primary functional entities responsible for executing business processes and serve as a major organizational backbone.
This marks a paradigm shift from previously introduced AI-augmented BPM systems~\cite{dumas2023ai} and other \emph{non-agent-centric} BPM approaches, where the term denotes that certain core entities (i.e., not agents) may exhibit some degree of agentic characteristic (e.g., autonomy), see~\cite{Fettke2025-vm}.

In an APM system, an \emph{agent} is an autonomous entity situated in an environment~\cite{wooldridge1999,Wooldridge2009}.
Functionally, an agent operates through a continuous control loop~\cite{DBLP:books/aw/RN2020}, alternating between \textit{sensing} its environment to update its internal state, \textit{reasoning} to select actions that align with its goals, and \textit{acting} to influence the environment.
An agent is \textit{proactive} and has its own thread of control, meaning that it makes (semi) autonomous decisions about which actions to perform and when~\cite{Vu25,LiZH25}.
In this sense, an agent processes information, generates content and knowledge, and proactively makes decisions and performs activities and tasks. This positions agents in contrast to classical components, objects, and services, which do not have goals and respond reactively to user prompts or inputs provided via an API~\cite{Wooldridge2009,weber2024,HassanLRGC00TOL24}.
For example, an agent may browse Web pages and make online purchases on behalf of a user; it may compare prices, select items, and complete checkouts~\cite{Gabriel25}.

The notion of an agent is sufficiently general to design environments in which heterogeneous entities perceive, reason, and act to achieve designated objectives. Within an agentic system, we distinguish the following categories of agents:
\begin{description}[wide=0\parindent]
     \item[Human agents,]\rmv{are agents whose embodiments are human beings}
     such as process workers or process managers.
     \item[Software agents~\cite{Wooldridge2009},] which inhabit a software environment and whose tasks are executed by means of program code, \emph{without deliberation}.
     \item[(Physically) Embodied agents~\cite{cassell2001embodied}] (a.k.a robots), whose decision-making and action directly affect the physical world.
     \item[AI agents,] \rv{that accomplish tasks} by means of deliberation via AI algorithms and models, for example, using generative AI techniques such as LLMs~\cite{DBLP:conf/ijcai/GuoCWCPCW024,DBLP:journals/corr/abs-2501-06322}.
\end{description}
\rmv{An illustrative example of an agentic system is a modern ride-sharing platform that connects a person needing a ride (i.e., a human agent) with an autonomous vehicle (i.e., an embodied agent) that provides the ride. Vehicle allocation and trip payments are handled by agents (AI and software, respectively).
The passenger agent has the goal of getting from point A to point B affordably and conveniently. The passenger also selects the service level (e.g., standard vs. comfort), whether to accept a suggested price, and whether to tip the autonomous vehicle (for suggesting the right music station and adapting the vehicle interior climate in response to passenger requests).
Vehicles' allocation to passengers is handled by an AI agent whose goal is to identify available vehicles at requested times and ensure resource allocation efficiency, assigning trip calls to vehicles at the nearest location while minimizing idle time for all vehicles.
The vehicle agent picks up trip calls within its zone of service and aims for an optimal navigation route to maximize earnings while minimizing time and costs.
Finally, upon trip completion, the payment is delegated to the payment processing agent that handles charging via credit card processing and issuing a digital invoice sent by email.} 

While an agentic system hosts a collection of inter-operating agents (i.e., it is a socio-technical system), it does not by itself realize any shared form of \emph{process awareness}. Here, we define process awareness as the assurance that agents’ inner workings conform to organizational processes and adhere to their operational constraints, regulations, and goals. Conceptually, we therefore regard an APM system as a system that explicitly realizes such process awareness in some concrete manner.

\vspace{.5em}\noindent Accordingly, in the context of BPM, we define an APM system as follows:

\begin{definition}
    An \emph{Agentic Business Process Management} (APM) system is a\rmv{n agentic} socio-technical system jointly realized by a collection of agents, some of which are at least partially \textbf{process-aware}\rmv{ agents}.
\end{definition}

\rmv{An APM system is an agentic socio-technical system, jointly realized by a collection of agents, some of which are at least partially process-aware.}

An APM system is considered socio-technical in the sense that it allows for a combination of social (human) and technical (technological) agents that interact and depend on each other to function effectively.
This does not exclude fully autonomous systems with little to no intervention by any human agent.
Collectively, agents execute business processes by virtue of each agent possessing a certain degree of process awareness. Thus, in an APM system, every agent is treated as \emph{autonomous}, with its own goals, decision-making capabilities, and knowledge of (parts of) the business processes. Agents may employ \emph{tools}, owned or shared, that reflect the objects, resources, functions, and services required to fulfill their goals.

We distinguish \emph{autonomy} from \emph{automation}. While automation denotes the execution of predefined tasks or workflows exactly as specified, APM systems facilitate autonomy, whereby agents can perceive, reason, and choose how to act within a process frame in order to achieve given goals. In other words, automation follows fixed rules, whereas autonomy in APM allows agents to make context-sensitive decisions while still respecting process awareness and constraints.

Revisiting our initial APM system example of the supplier onboarding process (Section~\ref{sec:intro}), we can see there are two types of process-aware agents: buyers and suppliers, some human and some AI-based.
The buyer agent is responsible for periodically disseminating Requests for Quotations (RFQs) to suppliers identified as suitable for meeting the manufacturing process targets and, in turn, evaluating the quotes they provide in order to decide which supplier to contract with.
The supplier agents work to fulfill their winning bids within individual time frames determined by the contractual terms. 
Each individual agent may also pursue other, non–process-aware goals that reflect the agent's specific role. For example, a supplier agent may be restricted to a particular geographical region. An APM system may also include agents that are not process-aware, such as a designated legal agent responsible for ensuring the diligence of the contracts.

We remain agnostic in this example about the specific mechanism used to instrument the agents with process awareness (e.g., how their individual knowledge and goals are set and updated), as there may be different ways of framing it within an APM system. For the purposes of defining the essence of an APM system, however, it is important that APM agents incorporate some concrete realization of such a framing capability (see Section~\ref{sec:capability-framing}).

Considering the agent as a first-class citizen in an APM system, we define an agent as follows:

\begin{definition}
An \emph{agent} in an APM system is a primary execution entity---an actor that perceives, reasons, and acts autonomously, with its autonomy framed to ensure process-aware behavior aimed at achieving process goals.
\end{definition}

\rv{We may call an agent in an APM system an \emph{APM agent}.}
It is important to note that\rmv{ in referring to an agent's own goals in the above definition,} once an agent becomes part of an APM system, its individual behavior and goals (i.e., micro-level work execution) are both framed and aligned.
The former implies that the agent's behavior is directed at both complying with process constraints (the \emph{frame}) and towards achieving the goals of the APM system processes (i.e., the macro level).
The latter entails that its behavior is harmonized with that of the other agents in the system. While being process-aware is an inherent property of an APM agent, it is a qualitative property in the sense that different agents in an APM system may vary in their degree of process awareness and autonomy, and in the extent to which their behavior helps achieve the system’s collective goals.

To realize this notion and equip an agent with the means to promote autonomy and process awareness in an APM system, we identify four essential capabilities the agent must possess: framing, explainability, conversational actionability, and self-modification.
These capabilities, and the rationale for their inclusion, are further detailed in Section~\ref{sec:capabilities}.

The actions of an agent are facilitated by the use of tools. We therefore define:

\begin{definition}
A \emph{tool} in an APM system is a means accessible to an agent that augments its capacity to reason, and to perceive and act upon its environment.
\end{definition}

Possible tools may include sensors, actuators, messaging, or other software functions and services. These tools may be accessed through communication protocols that define how information is exchanged.

In light of the presented concepts, next, we elaborate on the building blocks of an APM system and of an agent.

\subsection{Conceptual Architecture}
\label{sec:core}
\begin{figure}[tb!]
    \centering
    \includegraphics[width=1\linewidth]{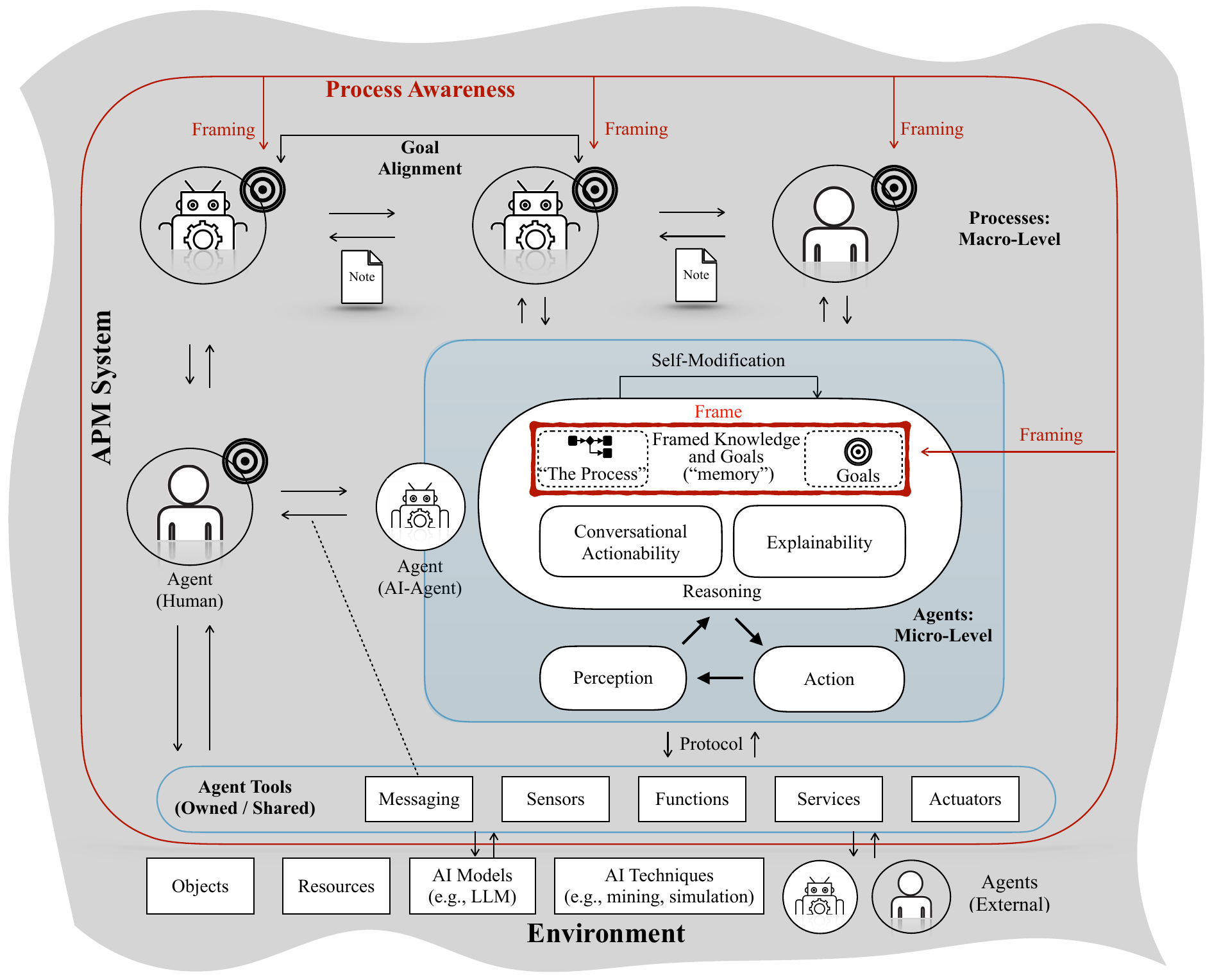}
    \caption{
    Conceptual architecture of an APM system depicting its key actors, components (boxes), and interactions (arrows). \emph{Process-awareness} (red-colored) is realized by a framing mechanism that constrains and aligns agents at the macro level, while at the micro level each process-aware agent runs a \emph{Perceive--Reason--Act} loop over framed knowledge and goals (its internal view of ``the process'' and its goals), as agents interact with other process-aware human- and AI-agents in the system, and with the external environment (external agents, objects, resources, AI models/techniques), using tools (messaging, sensors, services, and actuators).
    }
    \label{fig:apm_architecture}
\end{figure}

This section introduces the conceptual architecture of an APM system.
As visualized in Figure~\ref{fig:apm_architecture}, the architecture focuses on two distinct levels: the \emph{APM system} representing the \emph{macro level} management of work that establishes the process-awareness of its agents (Subsection~\ref{subsec:apms}), and the \emph{APM agents}, i.e., the autonomous execution of work in the APM system on the \emph{micro level} (Subsection~\ref{subsec:apma}).

\subsubsection{APM System}
\label{subsec:apms}

The APM system constitutes the ``world'' where agents live and act, and in which they can interact with each other and with the environment via tools (both individually owned and shared). These tools let the agents exchange data (e.g., \emph{notes}, \emph{messages}) with other agents in the APM system, employ a variety of AI models (e.g., LLMs) and techniques (e.g., simulation, mining), sense and manipulate external objects and resources, invoke remote functionalities via services, and interact with external agents that are not part of the APM system.
To understand each other, agents rely on a shared language, ontology, or protocol~\cite{Wooldridge2009}. Examples include Model Context Protocol (MCP),  Agent Communication Protocol (ACP), Resource Description Framework (RDF), and Knowledge Query and Manipulation Language (KQML).

The \emph{macro level} (process-aware management) represents the \emph{processes} themselves, defined by collective, organization-level goals. The attainment of these goals is achieved by imposing process awareness and goal alignment on the agents through a \emph{framing} mechanism.

\begin{definition}
    \emph{Framing} is a primary mechanism for ensuring process-awareness and goal alignment in an APM system, imposing restrictions on the autonomy of agents through their knowledge and goals. 
\end{definition}

\vspace{.5em}\noindent Without framed autonomy, an agent could perform any action in pursuit of its own individual goals. It provides the normative and operational specifications that agents must adhere to, including the \textit{lifecycle management} to create, suspend, or destroy agents as the process evolves.
More concretely, the realization of the framing mechanism attends to two aspects:

\vspace{.5em}\noindent\textbf{Process-Awareness} governs all agents’ work toward common process objectives. Process awareness entails that an agent’s autonomy is directed to ensure that its actions aim to fulfill the collective process goals. 

\vspace{.5em}\noindent\textbf{Goal-Alignment} sets the rules of engagement and structural relationships (e.g., hierarchies or coalitions), including the coordination of goals among agents that may work collaboratively. This also reflects the assignment of roles and segregation of duties throughout their lifecycle.

\medskip
The functional manifestation of the framing mechanism is expressed through the capabilities of the agents (see next section), while its concrete realization is an engineering choice, since there are various ways in which it can be implemented—for example, by assigning collective responsibility to one or more agents, through orchestration, or via a shared memory (e.g., as part of the frame) that allows agents to progress along a mutual plan.

\subsubsection{APM Agent}
\label{subsec:apma}

The \emph{micro level} (agent-oriented execution) consists of a collection of autonomous agents\rmv{, i.e., knowledgeable entities (human or software/embodied/AI-based) possessing individual goals. These agents} that jointly execute the process---and may participate in multiple processes.

Drawing on foundational ontologies like DOLCE~\cite{DBLP:journals/ao/BorgoFGGMPSV22} and GFO~\cite{DBLP:journals/ao/LoebeBH22}, agent-oriented modeling frameworks like \textit{i*}~\cite{Franch2024-ij} and Tropos~\cite{DBLP:conf/atal/GiunchigliaMP02}, and classic AI constructs such as BDI~\cite{DBLP:conf/icmas/RaoG95} and FIPA~\cite{foundations2003foundation}, as well as the mental model theory of reasoning~\cite{Johnson-Laird1991}, and in concert with the characteristics of an AI-Augmented BPM system~\cite{dumas2023ai}, we consolidate here a unified specification schema for an APM agent.

As conceptualized in Figure~\ref{fig:apm_architecture}, an agent, from an architectural point of view, consists of three main conceptual modules~\cite{Wooldridge2009}:

\vspace{.5em}\noindent The \textbf{perception module} allows the agent to observe the state of its \textit{environment}, \textit{context} (situational constraints, location), and other agents. 
These agents operate within the agent’s environment: some are part of the APM system (and thus process-aware), while others may be external to it.
This module is responsible for managing the agent's \textit{perception and sensing} attributes (e.g., sensors, percepts, belief update functions), ensuring the agent continuously updates its understanding of the world.
    
\vspace{.5em}\noindent The \textbf{reasoning module} is the core component for knowledge representation and decision making.
It processes sensory data and is continuously informed by the framing mechanism to maintain the agent's \textit{mental} (i.e., knowledge, perceptions, and beliefs) and \textit{intentional} (i.e, desires, intentions, and goals) models. This also includes the agent's ability to self-modify, learn, and explain its reasoning. The agent's core knowledge includes its \textit{identity} (i.e, agent ID, type), along with the agent's \textit{behavioral dispositions and plans} (policies, triggers, reactions) to decide what to do next. This module also inherently provides a computational mechanism for decision-making, enabling the agent to continuously process its knowledge, revise and generate ongoing inferences, and translate its perceptions into actions. Recent advances in AI explore the use of LLMs for this purpose, although their true capability to fulfill this role remains under debate~\cite{Saxena2024,khalid-etal-2025-large}. Several alternative approaches could support such a competence within an agent, including externalizing parts of the reasoning through specialized tools. Similar to the framing functionality, we intentionally leave the concrete realization of the decision-making component open.

The conceptual architecture introduces the following reasoning-related capabilities required for a generic agent to become \textit{process-aware} within an APM system:

\begin{description}[wide=0\parindent]
\item[Framing] (see~\ref{sec:capability-framing}).
This module represents the internalization of the APM system's \textit{framing}. Uniquely for an agent in an APM system, as a process-aware actor, the agent’s knowledge includes two key internal elements: a mental model and an intentional model. The internal mental model serves to represent the process reality, viewed by the agent as \emph{the process}, as perceived through the framing mechanism. It serves as the agent's ``memory'' regarding the process model, running instances, stakeholders, and all relevant process context. The internal intentional model maintains the agent's goals as they are shaped by its \textit{social} and \textit{normative} dispositions (i.e., obligations, roles, collective goals, and prohibitions). By explicitly consulting these frames, the reasoning component ensures that the agent's local constraints and objectives remain aligned with those of the overall process.

\item[Explainability]
(see~\ref{sec:capability-explainability}).
Supported by the agent's \textit{accountability} and \emph{traceability} attributes (i.e., trace logs, auditability), \emph{explainability} is the ability to articulate the rationale behind decisions. This ensures that process stakeholders can receive specific explanations of the workings of the socio-technical system.

\item[Conversational Actionability] (see~\ref{sec:capability-conversational}). 
Building on the agent's \emph{communication} and \emph{interaction} attributes (i.e., protocols, roles, languages), this capability allows the agent to negotiate and coordinate with other agents and act on the environment. It also enables process stakeholders to trigger actions, either at design- or run-time, to change system behavior. 

\item[Self-Modification] (see~\ref{sec:capability-self-modification}). 
Grounded in the agent's \textit{perception} abilities, self-modification is an agent's capability to adapt and evolve over time. Agents self-modify to better reach individual goals and, when successfully \emph{framed}, to achieve collective process goals.
\end{description}
    
\vspace{.5em}\noindent Finally, the \noindent\textbf{action module} allows the agent to perform actions that change the state of the environment and send messages to other agents. It defines the agent's \textit{capabilities} (i.e, skills, resources) and abilities to socially interact with other agents.

\section{Envisioned Capabilities of an APM system}
\label{sec:capabilities}
This section gives an overview of the aforementioned key capabilities of an APM system: \emph{framing}, \emph{explainability}, \emph{conversational actionability}, and \emph{self-modification}, as initially proposed in~\cite{calvanese2025autonomy,Fettke2025,Montali25,Senderovich25}.

\subsection{Framing in an APM system}
\label{sec:capability-framing}
The guard-railing of an autonomous agent requires assuring it is operating within its current \emph{frame}.
Intuitively, a frame is a set of rules, restrictions, and regulations, which may evolve over time. Frames establish boundaries within which one or several agents in an APM system may operate with maximal flexibility, making autonomous decisions~\cite{ACITELLI2025102573}. Frames may exist---at least---on \emph{agent type}, \emph{process}, and \emph{organization} levels (as well as potentially across organizations).


More analytically, frames are \emph{normative}: they specify deontic \rv{process} requirements \rmv{to the process}\rv{governing the behavior of the process}.
In contrast, classical process specification languages, such as BPMN and DECLARE, \rmv{which specify}\rv{are primarily concerned with specifying} behavior required to accomplish a business goal\footnote{BPMN is \emph{imperative}, specifying---at least \rmv{supposedly}\rv{ostensibly}---\rmv{exactly what needs to be done}\rv{the exact control flow of activities to be executed, and is therefore well-suited for \emph{operational} requirements.}\rmv{, while} DECLARE \rv{, in contrast,} is \emph{declarative}, specifying constraints \rmv{that need to be satisfied by otherwise flexible behaviors; still, both are \emph{operational}.}\rv{over permissible behavior}. \rv{While this declarative nature aligns naturally with \emph{normative} requirements, DECLARE is typically used to constrain execution rather than to explicitly capture deontic notions such as obligations, prohibitions, or permissions}.}. However, \rmv{in contrast to}\rv{unlike} informal definitions of business processes, e.g., as ``sets of activities'' performed to ``jointly realize a business goal''~\cite[p. 5]{weske2024business}, goals \rmv{are left}\rv{remain largely} implicit in these more formal and operationalizable process specification languages.

In previous works, operational specifications have been called \emph{frames} as well~\cite{dumas2023ai}.
Indeed, they can be considered a sort of \emph{operational} frame.
In APM systems ``frames'' focus, however, on the normative specification. When we need to distinguish, we call the two frames \emph{normative frame} and \emph{operational frame}, respectively. For example, an agent may be specified so that it must reject any student assignment submitted after the deadline. Such a requirement is considered a form of normative frame statement. An operational frame specification may complement this statement by instructing that, when rejecting an assignment, the agent shall: (1) retrieve the name and ID of the student who created the assignment, (2) record these details in the database, and (3) send the student an email informing them about the rejection. Alternatively, these three operational statements could be replaced by a single prohibition statement (i.e., a normative frame) not to delete the rejected assignment without informing the student who submitted it, leaving more room for autonomy in the behavior pursued by the agent, for example letting the agent decide on its own whether to first send an email to the student and then record the information in a database, or the other way around.

If there are no autonomous decision-makers, then the normative frame is just an additional condition over the operational frame; but if decision-making is possible, then the operational frame requires finding a strategy to satisfy the objective, whereas the normative frame requires choosing a strategy that remains within what is allowed (with respect to the frame).

Strategies for achieving goals under framed autonomy are associated with decision-makers, including software agents, giving rise to several problem setups, for centralized as well as distributed intelligence.

\vspace{.5em}\noindent\textbf{Centralized intelligence:}
    We consider the ``AI agents'' as a single entity orchestrating the \emph{process} that is executed in a mutually fully observable and coordinated manner. The \emph{environment} may be stochastic and not fully observable. The frame is over the process. The single entity may have active or passive responsibility for the frame. If we have multiple agents, we may break down the problem into several of the above scenarios.
    
\vspace{.5em}\noindent\textbf{Distributed intelligence:}
    We consider AI agents as distributed entities that enact the process as \emph{resources}. This has wide-ranging implications: a resource may have only partial observability of what other resources are doing; coordination may be effortful, and resource-level goals may be mutually inconsistent, or inconsistent with process-level goals.
    In such scenarios, we can frame individual resources, groups of resources, or the entire process.
    Accordingly, we need to assign responsibility to individual agents or groups thereof, and there may be strategic interactions affecting responsibility.

From these problem setups, we can derive three different blueprint scenarios for framed autonomy in business processes:
\begin{enumerate}[label=\itshape(\roman*)]
    \item we have a single decision-maker and place a frame on process behavior;
    \item we have multiple decision-makers and place frames on individual decision-makers;
    \item we have multiple decision-makers and place frame(s) on process behavior or parts thereof.
\end{enumerate}

In practice, there may be additional variance to the scenarios. For example, normative frames may be partially represented within operational process specifications, restricting overall agent autonomy. An example is a purchasing process where purchase orders can only be created and paid through a central IT system that enforces normative rules, e.g., regarding four-eyes approval policies.  Other parts of the global normative frame can potentially be projected to local agent-level norms. For example, overall spending limits may apply on the global level, but could be operationalized locally.
\rv{Specifically, local operationalization of frames acts as a security and privacy boundary. Indeed, the frame limits an agent’s perspective and capabilities to only its necessary process context, inherently restricting the potential impact if an agent is compromised or manipulated by a deliberately malicious actor.}

\subsection{Explainability in an APM system}
\label{sec:capability-explainability}
As a prerequisite to effective execution and governance, an APM system should be explainable. Because an APM system is a composition of AI agents, this may be achieved by endowing agents with the intrinsic capability to explain their own behavior and actions, including those agents assigned collective process-awareness responsibilities. To this end, each agent’s specification should include explicit instructions concerning the requirement to articulate the explanans (the explanation itself) in response to certain triggering situations or conditions (explanandum), as well as which explanation mechanisms---so-called \emph{eXplainable AI} (XAI) techniques---may be employed.

Empowering agents in APM systems with explainability will help address important concerns, including the following:

\begin{description}[wide=0\parindent]
    \item[Trust] issues may arise among stakeholders---including process owners, business analysts, end users, and customers---who may hesitate to rely on agentic process recommendations or automated agent decisions when the underlying rationale is unclear.
    \rmv{The opacity of the inner workings of agents employing AI (for their behavior and reasoning) may make it difficult to \textit{debug}, identify potential failures, or understand why an agent performed a certain action or task.}   
    \item[Accountability] for agent behavior requires that if an agent fails, it can explain the underlying rationale such that responsibility can be assigned and corrective actions can be implemented.
    \item[Biases] may be perpetuated by AI and ML components underlying APM systems and their agents. Such biases may lead to discriminatory or unfair agent decisions; explainability is a prerequisite for detection and mitigation.
    \item[Compliance] of agents' behavior with regulatory frameworks, such as the EU's GDPR and AI Act, needs to be demonstrable. This requires an increasing level of transparency, particularly in high-risk domains like finance, healthcare, and human resources, which are common areas for BPM applications.
\end{description}

\rmv{Unlike traditional BPM or Robotic Process Automation (RPA) systems that follow rigid, predefined rules and workflows, the degree of autonomy in APM systems implies.}
\rv{Agents in APM systems make 
independent decisions, adapt to changing conditions, and learn from experience} with minimal human intervention. Here, explainability offers a central mechanism through which agents can articulate the rationale behind their behavior. 
As such, explainability becomes a first-class citizen in the realization of APM systems, supporting agent autonomy from two perspectives: \emph{(i)} enabling agents to independently resolve misalignment in other agents' behavior; \emph{(ii)} reducing human intervention by making agent behavior understandable and transparent.

To account for the quality of explanations in APM systems (e.g., soundness, accuracy, usefulness, and interpretability), derived from situation-aware explainability ~\cite{Amit2022,DBLP:conf/bpm/AmitFLS22,Fahland2024} and causal processes~\cite{Fournier2023v3,Yuval2026}, we postulate that the following are desirable properties about explainability in APM systems:

\begin{enumerate}
    \item Ability to infer explanations that conform to the framing constraints and statements.
    \item Ability to capture the richness of contextual information that affects agents' behavior and decisions.
    \item Ability to reflect causal execution dependencies among actions in the agents' trajectories/behavior.
    \item Ability to provide explanations that are interpretable to other agents (humans or digital).
\end{enumerate}

\subsection{Conversational Actionability in an APM System}
\label{sec:capability-conversational}

As the term suggests, conversational actionability refers to an agent’s ability to combine interaction and enactment capabilities. This is essential in an APM system. On the one hand, agents need to integrate conversational capabilities to coordinate with one another and to receive instructions from, interact with, and report to the human agents involved in `the process' in various roles (ranging from active participation to management). Knowing what the process is and who the agents assigned to relevant roles are may be considered part of an agent’s mental model of the process. On the other hand, agents must be able to link such conversations to the corresponding enactment capabilities, making decisions and performing actions accordingly.

In \cite{Montali25}, conversational actionability is positioned within AI-augmented BPM Systems as the ability of the system to handle these two requirements:
\begin{description}[wide=0\parindent]
\item[Process-aware conversation.] The system can interact with users or external agents using a conversational interface to support, trigger, and guide actions related to the enactment of one or multiple processes. 
\item[Process-aware actionability.] Conversations trigger business process executions, such as taking a decision or performing an action, as well as evolution actions, such as deciding on priorities or changes to the process.  
\end{description}

\rmv{When moving from a centralized execution system to an APM system, these capabilities must be collectively realized by the set of agents operating within the APM system, also deciding how each agent instantiates these two capabilities based on its own individually informed perspective about `the process' (i.e., its frame).}
\rv{When moving from a centralized execution system to an APM system, these capabilities must be collectively realized by the set of agents operating within the system. Instead of relying on centralized intelligence and a single point of truth, an APM system caters to federated perspectives. Each agent decides how it instantiates these two capabilities based on its own individually informed perspective about `the process' (i.e., its frame). Hence, this may require the use of consensus resolution approaches or the ability of external agents to tolerate a possible variety among the results when interacting with different agents.}

As for the conversation component, an agent may, in fact, interact with other agents using natural language and/or exploiting other unstructured and semi-structured information sources (such as diagrams and charts) or formal communication languages, such as multi-agent interaction protocols. Choosing which interaction modalities are embodied by the agent, manifested by the proper selection of tools in the APM system, obviously also depends on whether an agent only interacts with other computational agents, and/or humans. 

In terms of actionability, depending on the role played by the agent, it may need to cover one or more key functionalities (i.e., types of agent-enacted behavior; cf.~\cite{Montali25}):\vspace{-10pt}

\begin{description}[wide=0\parindent]
\item[\emph{Query} --] to provide information on the process(es), either at the model level or regarding execution data pertaining to the current or past states of affairs;\vspace{-7.5pt}
\item[\emph{Recommend} --] to provide insights and suggestions on the adaptation and future evolution of process instances;\vspace{-7.5pt}
\item[\emph{Create} --] to elicit models from domain knowledge and process-relevant data;\vspace{-7.5pt}
\item[\emph{Execute} --] triggering actions to move process instances to a new state.
\end{description}
In order to enact such behaviors, agents typically require interacting with tools providing different services or the same service with different functional and non-functional guarantees. For example, determining the likely time-to-completion for an order may be achieved using predictive monitoring techniques, employing simulation, or by a combination of the two.
This creates the challenge of identifying the best mix of tools to realize an overall functionality in the ``best way'' possible (see Section~\ref{sec:action}).

\rmv{As opposed to traditional BPM systems, which rely on centralized intelligence and a single point of truth for the realization of such functionalities, an APM system caters to federated perspectives. This may require the use of consensus resolution approaches or the ability of external agents to tolerate a possible variety among the results when interacting with different agents.}

\subsection{Self-Modification in an APM System}
\label{sec:capability-self-modification}
In order to adjust to ephemeral or permanent changes, APM systems must be able to self-modify.
A fundamental distinction in self-modifying APM systems is between \textit{adaptation} and \textit{evolution} of both individual agents and agent coalitions. Adaptation and evolution fundamentally differ in scope, duration, permanence, and the type of knowledge they leverage.

    \vspace{.5em}\noindent\textbf{Adaptation}
    refers to short-term, instance-specific modifications~\cite{weyns2020,PalmMP20} that address immediate, unforeseen issues during process execution. These modifications do not alter the underlying process model or schema, but rather constitute localized modifications performed by the agent within the constraints of the current process definition to handle exceptional circumstances or environmental changes that were not anticipated at design time. Adaptations are ephemeral---they affect only the current process instance and do not propagate to future executions unless explicitly learned and incorporated into the model (which then would be considered evolution). Such adaptations can be enacted individually by an agent, or they can be enacted collaboratively by engaging with other agents about how their shared perspective on the process may need to be adapted. Adaptations may be reactive (i.e., in response to perceived deviations or problems) or they may be proactive (i.e., performed based on predictions), see~\cite{Senderovich25,MetzgerKRP23}.
    
    \vspace{.5em}\noindent\textbf{Evolution}
    describes long-term modifications to the process logic, model, or policy that persist across multiple instances and executions. These longer-term modifications are typically shared among multiple agents in the APM system. Evolution may be informed by aggregated insights, patterns, and correlations discovered through analysis and learning from historical execution data, performance metrics, and repeated anomalies. Rather than responding to a single event, evolution synthesizes knowledge from multiple instances to identify systemic issues, improvement opportunities, or changing environmental conditions that warrant permanent changes to how the process is defined or executed. These modifications affect the process model itself and thereby influence its future instantiations, i.e., future processes \cite{DBLP:journals/dke/RinderleRD04}. Such evolution also entails the need to communicate and re-align how it is applied across all relevant agents in the system. \\

Ideally, the relationship between adaptation and evolution forms a feedback cycle in the APM system. 
Adaptations generate execution traces and performance data that feed into agents' learning mechanisms. 
When certain adaptive patterns prove consistently effective across multiple instances or contexts, they become candidates for evolutionary incorporation into the process model~\cite{DBLP:journals/ijcis/WeberRRW09}, and hence into the agent's frame.
Conversely, evolution should reduce the frequency of certain types of adaptations by preemptively addressing known failure modes or inefficiencies. 
However, adaptations remain necessary for handling truly novel situations that have not been anticipated during design time~\cite{Unexpected-2020} or not yet encountered frequently enough by any agent in the APM system to justify evolutionary modifications, or to address context-specific conditions that should not be generalized.

The distinction between adaptation and evolution has important implications for system design. Adaptation mechanisms must prioritize responsiveness, robustness, and security/safety under uncertainty, often operating with incomplete information and limited time for deliberation. 
Adaptation mechanisms require runtime monitoring, exception handling capabilities, and flexible execution engines that allow deviating from prescribed process execution paths. 
Evolution mechanisms, meanwhile, require sophisticated analytical capabilities: pattern mining across execution histories, causal inference to distinguish correlation from causation, statistical validation to ensure that observed patterns are not artifacts of noise or bias, and change management protocols to securely/safely deploy model modifications without disrupting ongoing operations.

\section{Challenges}
\label{sec:challenges}
This section provides an overview of research challenges that require solving to further advance the relevance of APM.
The challenges are categorized according to the four core capabilities of \emph{framed autonomy}, \emph{explainability}, \emph{conversational actionability}, and \emph{self-modification}. Analogously to the capabilities in the previous section, these challenges were identified during the Dagstuhl seminar and published as PMAI workshop papers at ECAI~\cite{calvanese2025autonomy,Fettke2025,Montali25,Senderovich25}. We augmented and refined these challenges based on the instrumental feedback received from the PMAI workshop participants.

In addition, we introduce a set of \emph{cross-cutting} challenges that affect several of the aforementioned capabilities.

\subsection{Challenges Regarding Framed Autonomy}
\label{sec:formal}

Achieving framed autonomy in APM systems comes with practical challenges, ranging from fundamental questions about the notion of an agent in business processes to specification and operationalization.

\paragraph{F1: What is a pragmatic notion of an agent in the context of business process execution?}
    Before the broad adoption of LLMs, the notion of an agent did not play a major role in the engineering of business information systems and the processes that run them. Consequently, practitioners cannot be expected to be familiar with the depth and sophistication of agent-related abstractions and the presented framing mechanism for the realization of process awareness. To the contrary, a practitioner may consider a software tool that makes use of an LLM as an agent, without much thought about further properties. Defining a more precise and robust notion of an agent that is still intuitively understandable by business process management practitioners can thus be considered a key prerequisite for framing agents' autonomy.

\paragraph{F2: How to elicit and specify frames?}
    The elicitation and specification of mental and intentional frames requires a \emph{frame meta-model}, and one or several specification languages. To this end, existing specification languages can be reused; potentially, several languages and their underlying concepts can be combined. For example, declarative approaches to process specification---such as \declare~\cite{DBLP:books/sp/22/CiccioM22} and in more practical contexts business rule and query languages with temporal reasoning capabilities~\cite{DBLP:conf/bpm/KampikO24}---can be augmented with deontic notions to promote normativity to a first-class abstraction.
    For elicitation, both symbolic and subsymbolic approaches can be used and fused.
    LLMs can generate frames or parts thereof from natural language text, whereas rule mining approaches can be applied to infer normative constraints from the traces of well-behaved agents and multi-agent systems.

\paragraph{F3: How to operationalize frames on real-world symbolic data?}
    Once specified, frames need to be integrated with APM systems to ensure the agents’ frame-compliance during runtime. A short- to mid-term prerequisite is the operationalization of frames using technologies that do, in fact, run in large organizations. Here, explainability is a necessity, considering the practical intricacy of normative requirements, as well as the scale of real-world symbolic queries and data. To address this runtime compliance, formal logic-based approaches~\cite{CASCIANI2026102718} can provide the imperative logic foundation needed to enforce the operational boundary of the agent. By leveraging strict transition semantics, such formalizations can naturally guardrail the agent's execution to legal actions and safely handle exceptions via guarded execution constructs.

\paragraph{F4: How to design incentive and reinforcement mechanisms for frame-compliant agents?}
     
     A central challenge in APM is understanding how to design agents whose reasoning and behavior consistently align with their prescribed frame, meaning the norms, goals, constraints, and process logic meant to guide their actions. The question is not merely how to make such a frame available to agents, but how to ensure that agents actively use it, internalize it, and treat it as instrumentally valuable. This requires developing ways to define, measure, and strengthen an agent’s degree of frame compliance as a continuous, learnable property. A prerequisite are incentive structures that cultivate an agent’s ``desire'' to be process-aware. Humans can be motivated through monetary compensation, perks, recognition, or intrinsic cultural alignment. AI agents require a functional counterpart that reshapes their internal utility models. The open problem is how to design rewards inherently as part of the frame, that make adherence to the frame beneficial for the agent while still preserving autonomy. To support this, reinforcement mechanisms must be woven directly into the agentic lifecycle. Over time, agents must learn which components of the frame are genuinely useful, which elements constrain them productively, and which parts might be deprioritized without violating safety, governance, or organizational policies.

\subsection{Challenges Regarding Explainability}
\label{sec:explainability}

Challenges with respect to the explainability in APM systems center around technically sound explanations that are easily understood and actionable, given their intended and actual explainees.

\paragraph{X1: How to specify or let agents learn other agents' preferences regarding explanations?}
    A variety of elicitation mechanisms may be used in an APM system to effectively capture explanation preferences through various channels, whether explicitly declared upfront in the agentic specification, interactively elicited through dialogue, or implicitly inferred from user behavior.

    Systems must accommodate both static preferences that remain consistent and those that dynamically adapt to changing contexts, while supporting the natural evolution of preferences as the explainee's, human or digital, understanding develops. Here, the challenge is to find the sweet spot between keeping explanations up to date and not confusing the explainee (i.e., limiting its autonomy). Finally, we have to think about when and how to provide full versus incremental explanation updates. 

\paragraph{X2: How to leverage and extend existing XAI techniques by the agents to generate explanations?
}
    Employing state-of-the-art XAI techniques for APM systems has several limitations. This is because making sense of agent behavior goes beyond explaining the AI models used under the hood; it must also account for the broader context in which agents perform their actions, the knowledge they may have acquired over time, possible cause-and-effect relationships among their actions, and the constraints (i.e., framing) imposed on their behavior.
    We therefore need new and enhanced explainability techniques that help agents explain the outcomes of their behaviors and decisions more broadly.
    \rmv{Examples include what-if analyses and process outcome analyses. 
    In the future, it should be clarified how existing techniques can be integrated.}

\paragraph{X3: How may one articulate actionable explanations (e.g., to other agents) to preserve autonomy? 
}
    \rv{Besides being context-sensitive, explanations should be actionable}---indicating to the explainee which corrective or mitigating actions could be taken to alter the state of the explanandum, particularly without escalating the situation to any external agent. In this way, the explainee may be able to autonomously act upon the condition at hand. However, further work is needed to devise a systematic approach that enables the explainer to determine the most effective content for the explainee, to elicit such corrective action—taking into account both the explanandum and the behavioral intentions of the explainee. 
    
\paragraph{X4: When to generate explanations (generation time) and how long to preserve them?
}
    With respect to APM system performance and memory efficiency, it is important to assess whether explanations should be generated proactively wherever feasible, or whether their generation should be deferred until necessary. A related question concerns when outdated explanations ought to be retired~\cite{mehdiyev-2021}.

\paragraph{X5: How to generate causally sound explanations?
}
Not every pair of action executions is causally dependent. To generate sound explanations, it is therefore important to distinguish spuriously correlated actions (e.g., those that are merely temporally sequential) from causally dependent ones (i.e., when one action directly triggers the execution of another) within an agent’s behavior or across multiple agents. 

\subsection{Challenges Regarding Conversational Actionability}
\label{sec:action}
Challenges in conversational actionability pertain to how agents can realize a virtual circle of action and interaction with human principals such as domain experts and other agents in order to enact the process faithfully and efficiently. 

\paragraph{A1: How to engage in conversations with human principals and other agents?} The first challenge pertains the \emph{conversational} facet of an APM system. This contains two distinct aspects. The first concerns the need for agents to interact with human principals--typically using natural language or domain-specific languages (such as process diagrams, dashboards, and the like). As for natural language in general, there are only few studies focusing on the ability of foundation models to converse about processes, with many open challenges discussed in a dedicated vision paper~\cite{largeprocess}, and some preliminary contributions~\cite{NeZG25,DBLP:journals/jiis/BernardiCCM24}. This is even more challenging when natural language conversations are mixed with domain-specific knowledge sources. 

The second aspect pertains to agent-to-agent interactions and data exchange, which can be carried out using a variety of different means, from specific interaction protocols and recently developed human-readable token encoding formats (e.g., TOON\footnote{\url{https://github.com/toon-format/spec}}), and formal languages to more flexible forms (e.g., based on higher-level, informal languages).

\paragraph{A2: How to exploit capabilities of process management tools and services?} The second challenge pertains to the \emph{actionability} facet of an APM system. Agents need to relate conversations to goals and, in turn, to actions. This imperative raises a twofold challenge. 

On the one hand, the actions in question are not under the direct responsibility of the agents, but require them to interact with different \emph{actuators}, such as: (1) ERP, CRM, or other systems of record to perform transactions; (2) collaboration tools, e.g., to trigger notifications to human actors; 
(3) physical actuators, e.g., IoT devices or robots; (4) process or task automation tools to update the state of a process or to trigger predefined automation scripts; (5) dedicated planners and schedulers to trigger sequences of steps to achieve a desired state.

On the other hand, in deciding which action should be selected, agents typically need to acquire relevant information about the organisation and, in particular, the current state of affairs as part of their individual mental model about the process. This can only be done if the agents can effectively exploit tools and services encapsulating specific, well-understood, and highly trusted functionality related to data and process intelligence, covering all phases of the process lifecycle (from model-driven analysis and simulation to online data querying, process mining components, as well as predictive and prescriptive components). This calls for encapsulating such components as tools into semantically rich descriptions that can be used by agents, describing functional and non-functional requirements and enabling discovery (for example via MCP).

\rmv{\paragraph{A3: How can an AI-augmented business process management system expose itself as an agent?}
An organization adopting the paradigm of an AI-augmented business process management system (ABPMS)\cite{dumas2023ai} may choose a mixed approach to modernizing its existing legacy BPM system by embedding AI augmentation into some of its process functionalities, thereby introducing intrinsic elements of (framed) autonomy that allow certain decisions to be made without, or with only partial, direct human oversight. In such settings, the entire ABPMS itself can be wrapped by one or more agents for overall process oversight. This raises the question of how such a wrapping agent interacts with the inner components of the existing system for actual process enactment.}

\paragraph{A3: How to compute and use key indicators on the overall behavior?}
A key challenge related to conversational actionability is how to compute and use key functional and non-functional process indicators obtained when gathering information and taking actions. Such indicators range from standard KPIs related to performance, time, and cost, to a larger set of indicators measuring/estimating trust, usability, uncertainty, flexibility, and alike (see~\cite{Montali25}). A set of KPIs to assess completeness and correctness of process models when compared to corresponding textual descriptions have been proposed~\cite{DBLP:conf/bpm/KlievtsovaBKMR23}) and vice versa for generated text from process models in~\cite{DBLP:conf/re/KlievtsovaMKR24}. Moreover, for example, an agent may need to ponder the impact of taking an action and, in doing so, may invoke a simulator and/or a black-box predictor. Each of these components would provide different levels of trust, precision, and uncertainty, and depending on how they are used (e.g., selecting one of them or selecting both and then aggregating the obtained outputs), such levels would propagate up in completely different ways. This is essential to obtain end-to-end indicators regarding the overall process execution.

\paragraph{A4: How to balance autonomy, delegation, and control?}
The presence of multiple, interacting agents, operating using different tools and AI models with different levels of trust, performance, cost, etc, makes it even more challenging to determine how to balance autonomy and human oversight, and how to achieve the best trade-off between delegation (improving performance) and control (mitigating undesired outcomes and retaining human responsibility).

\subsection{Challenges Regarding Self-Modification} 
\label{sec:modification}

Below, we present the relevant challenges related to the capability of APM systems to adapt and evolve.

\paragraph{M1: How to govern self-modification?}
For the agents in an APM system to operate safely and effectively, governance and human oversight must be built into their design. A key question is when to refrain from autonomy and return control to a human agent, e.g., in scenarios where the autonomous agent has insufficient confidence and is at risk of making mistakes.
Research in AI planning, runtime monitoring may help establish thresholds or confidence bounds beyond which a process must escalate to human decision-makers. Techniques from the ML literature on prediction with reject option~\cite{hendrickx2024machine} (sometimes called ``learning to defer'') can be explored as a solution direction.

Similarly, determining when and how a human agent should validate a proposed process or policy modification requires a framework for explainable adaptation, where the autonomous agents present justifications for their suggested modifications. Solution directions may include methods from the field of explainable AI (XAI)~\cite{dwivedi2023explainable} (see~\ref{sec:capability-explainability}), or justifications may be provided in alternative forms such as simulations of expected outcomes.

Another major issue is aligning the planning and goals of the AI agents in the system with those of its human agents (i.e., process stakeholders)~\cite{Sreedharan2022-jp}. Multi-objective optimization techniques~\cite{marler2004survey} can assist in balancing trade-offs between performance, cost, compliance, and user satisfaction while maintaining logical constraints defined in the process model. However, optimizing across conflicting objectives remains a challenging problem, especially when human values or other non-quantifiable criteria are involved.

\paragraph{M2: How to evaluate the success or failure of modifications?}
Quality assurance in fully autonomous scenarios introduces its own challenges. Without humans routinely checking outcomes, the agents in the APM systems must develop internal mechanisms for evaluating whether their adaptations ``worked''. Here, ML-based anomaly detection, performance baselining, and causal reasoning can help detect regressions or misbehavior. Formal methods for verification and validation of modifying process logic or agent policies also present a relevant research direction here. Generative AI, particularly LLMs, could also be used for generating justifications or summaries of decisions for human audits or by providing labels for downstream evaluation---an emerging area sometimes called LLM-as-a-judge~\cite{zheng2023judging}.
Solutions to the quality assurance problem could use human input purely for evaluation, even in scenarios where the execution is fully automated.
However, human input is costly, and hence a challenge is to make this cost-efficient with maximal autonomy. Directions to be explored could include active testing~\cite{kossen2021active} or online testing~\cite{SammodiMFOMP11}.

\paragraph{M3: How to align concurrent evolution and adaptation of linked process executions among process-aware agents?}
Business processes rarely exist in isolation. In practice, APM agents operate within ecosystems of interconnected process executions, shared resources, and distributed agent coalitions that may themselves be undergoing evolution and adaptation. 
For example, agents engaged in related processes may undergo similar adaptations and thus may be able to share ``lessons learned'' from these common adaptations, thereby making the overall modification more effective (e.g., see~\cite{DraganEtAl2025}, where this is explored for self-adaptive software systems).
A key challenge is detecting and managing these interdependencies in real-time. APM agents must be aware not only of their own mental and intentional states (i.e., processes and goals), but also of the adaptation state of related agents engaged in the same processes. This requires mechanisms for inter-process communication, coordination protocols, and potentially a meta-level orchestration agent that monitors system-wide adaptations. Techniques from distributed systems, such as consensus algorithms, distributed constraint optimization, and multi-agent coordination~\cite{shoham2008multiagent}, may provide foundational approaches. However, these must be adapted to the dynamic, semi-autonomous nature of APM systems, where agents might not have complete visibility into other agents' internal states or future intentions.

\paragraph{M4: How to enable continuous learning and adaptation management?}
A defining feature of self-modifying agents in APM systems is their ability to learn from experience and improve process behavior over time. This may be achieved utilizing process mining techniques by the agents themselves as a self-reflection mechanism to analyze their own recorded behaviors (a.k.a., agent trajectories). To support this, the agents must continuously record the adaptations they make and assess their impact both locally (per agent) and globally (across agents). Capturing this meta-knowledge creates a feedback loop where successful modifications reinforce future decisions and ineffective ones are pruned. AI planning and reinforcement learning techniques (while balancing exploration and exploitation; e.g., see~\cite{PalmMP20,satyal2019business}) are promising approaches to structuring such learning loops, especially when context-aware and enriched by causal modeling of intervention outcomes.

However, scaling these techniques may pose challenges:
Long-running processes or high-volume APM systems face constraints of bounded memory and context. The agents in the system cannot retain or process each event they have ever observed. Hence, the agents must learn to construct and maintain bounded knowledge representations: compressed summaries, predictive state abstractions, or fixed-size windows of relevant execution history. Techniques from stream reasoning, process mining over sliding windows, and transformer-based sequence models may offer solutions. Designing APM agents that know which parts of their history to remember, forget, or query becomes a core challenge for sustainable, real-time, continuous learning.

\paragraph{M5: How to model and measure uncertainty in APM systems?}
Non-determinism in agentic behavior may arise in APM systems due to \emph{(i)} stochastic AI techniques and models (such as LLMs); \emph{(ii)} stochastic and/or drifting contexts; \emph{(iii)} the difficult-to-anticipate actions of human agents.
The agents in APM systems thus must operate under both aleatoric (inherent randomness) and epistemic (lack of knowledge) uncertainty~\cite{hullermeier2021aleatoric}. \rmv{Addressing these uncertainties requires both awareness and expressiveness.}
It is well-known from the literature on uncertainty quantification and active learning that accurate estimates of specifically \emph{epistemic} uncertainty are a prerequisite to learn and adapt policies to new environments in a data-efficient way~\cite{hullermeier2021aleatoric,nguyen2022measure,tharwat2023survey}.

Techniques from probabilistic modeling, Bayesian inference, and fuzzy logic are key for representing and reasoning about agentic uncertainty in process executions.
In some domains, thresholds or confidence levels may be set by human experts (qualitative), while in others, Bayesian models or statistical metrics (quantitative) provide actionable measures. As an example, reliability estimates for proactive adaptation may be derived from ensembles of prediction models~\cite{MetzgerKP20}.
APM agents must combine these forms of knowledge, learning from past executions when quantitative models are feasible, while falling back on qualitative heuristics when data is sparse or ambiguous. Hybrid methods that integrate fuzzy logic or ensemble learning could further improve uncertainty modeling in domains with imprecise information.
Deciding when to defer the decision-making back to a human agent is a classic task where most solutions involve uncertainty quantification~\cite{hendrickx2024machine}.

\subsection{Cross-cutting and Complementary Challenges}
\label{sec:crosscutting}

\paragraph{C1: Toward Automated Provisioning and Onboarding of Legacy BPM into APM.}
    \rmv{A key challenge in the evolution from traditional BPM to APM lies in the provisioning and onboarding of existing process assets, workflows, and operational knowledge into an agent-driven system.}
    \rv{A key challenge in this direction lies in provisioning and onboarding legacy process assets into an APM system.}
    Organizations often operate long-standing BPM and workflow systems, implemented across diverse technologies and custom solutions.\rmv{—that, while still functionally adequate, increasingly suffer from technological obsolescence, limited autonomy, and substantial reliance on human intervention.} Many such systems require extensive manual effort for data handling, exception management, and cross-system orchestration, which in turn introduces operational inefficiencies and error risks. Moreover, critical business logic is frequently fragmented across workflow diagrams, scripts, rules, and tacit human practices, making structured extraction and reinterpretation difficult. Transitioning to APM therefore demands not only technical migration but also semantic and operational transformation, namely \emph{agent-centric process mining}, including the reinterpretation of workflows into agent-oriented roles, goals, policies, and behaviors, while preserving institutional knowledge and ensuring continuity of governance and auditability. How this process can be aided, automated, deployed, and validated may keep the industry busy for years to come, as organizations seek reliable methodologies and tooling to safely and incrementally modernize their automation capabilities.

\paragraph{C2: Holistic Security and Privacy in Agentic Workflows.}
    \rmv{APM systems introduce a unique attack surface that traditional BPM security models (mainly based on role-based access control) cannot fully address.}
    \rv{APM systems introduce a unique attack surface, requiring novel security models (mainly based on role-based access control).}
    The capability of \emph{Conversational Actionability} implies that agents may ingest malicious prompts (prompt injection) or interact with compromised external agents, potentially leading to data exfiltration or unauthorized actions. 
    \rv{Simultaneously, \emph{Self-Modification} capabilities introduce the risk of ``poisoning'' agents' memories}~\cite{torra2026memorypoisoningsecuremultiagent} to ultimately cause them to adopt unsafe behavioral patterns. 
    Furthermore, while \emph{explainability} is required for trust, it presents a privacy paradox: detailed explanations of agent reasoning may inadvertently leak sensitive business data or personally identifiable information to unauthorized third parties. 
    \rmv{A major cross-cutting challenge is, therefore, the development of security frameworks that can govern the trade-off between agent autonomy and information security, ensuring that agents can negotiate and adapt without compromising the integrity and confidentiality of the process-aware environment.}\rv{To prevent deliberately malicious actors from modifying workflows, APM systems must adopt advanced defense mechanisms. Technical and semantic guardrails (such as gateways and strict input/output sanitization) can intercept malicious instructions, and more advanced architectural design patterns such as ``Action-Selector'' or ``Plan-Then-Execute'' can constrain agents to predefined API calls, preventing them from executing unauthorized workflow modifications~\cite{DBLP:journals/corr/abs-2506-08837}. Moreover, process mining techniques can be repurposed as real-time security monitors to automatically detect potentially malicious deviations in process execution~\cite{DBLP:conf/caise/Cobo-ArizaAQVG25}. However, genAI agent security is an emerging and evolving area of research (cf.~\cite{11447227} for a taxonomic overview) and, accordingly, APM systems face---just like any other systems that deploy genAI agents---many security challenges that remain unsolved.
    An example of an open challenge is ``indirect prompt injection''~\cite{DBLP:conf/emnlp/WangSYSNZWGS25}, where malicious instructions are hidden within external data or documents the agent is tasked to process.
    Thus, a major cross-cutting challenge is the development of security frameworks that can govern the trade-off between agent autonomy and information security. As overly rigid guard-railing limits adaptive capabilities, it is important to find a balance that ensures agents can negotiate and adapt without compromising the integrity and confidentiality of the process-aware environment.}

\paragraph{C3: Benchmarking and Evaluation Frameworks for APM.}
    While individual capabilities of AI models (e.g., LLM reasoning) are frequently benchmarked, there is a distinct lack of holistic evaluation frameworks for agentic systems in a process management context. 
    \rmv{Traditional BPM characteristics (such as time, cost, and quality) are insufficient to comprehensively characterize agent qualities.} 
    In addition \rv{to well-established BPM characteristics, such as time, cost, and quality}, we require novel metrics to assess how well the agents in APM systems deliver their capabilities (see Section~\ref{sec:capabilities}).
    While this also calls for novel benchmark datasets to evaluate how well agents adhere to these qualities in stochastic environments, the benchmark datasets may be compromised by \textit{data pollution} (aka. \textit{data contamination} or \textit{benchmark contamination})~\cite{baltes2025,zhou2023}.
    This occurs when an AI model, used as a tool in the APM system, has been accidentally or intentionally exposed to the benchmark data during its training or fine-tuning phase. This has several negative consequences: \textit{(a)} The AI model's performance will be artificially inflated, giving a false sense of its capabilities; \textit{(b)} the purpose of a benchmark is to objectively measure performance on unseen data, yet contamination undermines this, making it difficult to reliably perform comparisons; \textit{(c)} the AI agents employing such a tool that excels on a contaminated benchmark are likely to perform poorly in real-world applications where they encounters truly novel data.    
    These are very important challenges to overcome to move APM from theoretical viability to industrial adoption, allowing organizations to quantify the return on investment (ROI) and risk profile of deploying autonomous workforce agents.

\paragraph{C4: Liability and Accountability.}
    The shift from human-centric to agentic process execution creates a ``responsibility gap'' that cuts across legal and organizational dimensions. 
    When the agents in an APM system operate within \emph{framed autonomy}, determining liability for adverse outcomes becomes complex: is the error attributable to the software developer, the organization enforcing the regulations (frame) onto the APM system, the specific AI model provider, or the emergent behavior of the multi-agent system? 
    This challenge is exacerbated by \emph{self-modification}, where an agent's behavior may drift significantly from its initial design. 
    Research must align APM system architectures with emerging legal frameworks (such as the EU AI Act) to establish clear chains of accountability. 
    This involves not only technical traceability but also the development of ``social'' contracts between human and digital agents, defining the boundaries of delegation and the precise conditions under which human intervention is legally and operationally mandatory.

\paragraph{C5: Methods for Engineering APM Systems.}
    One complementary challenge is to build on the aforementioned challenges for establishing essential architectural and technical building blocks of agentic process management and deliver novel \textit{analysis and design methods} to support the engineering of APM systems\footnote{e.g., see the CAiSE 2026 workshop \href{https://sites.google.com/view/epais2026}{EPAIS}.}.
    Such engineering methods may build on agent-oriented analysis and design methods, such as AAII, Gaia, Tropos, or Prometheus~\cite{Wooldridge2009}.

\section{Conclusions and Outlook}
\label{sec:discussion}

Our work extends the BPM community's well-established tradition of writing manifestos.
Notably, the \emph{Process Mining Manifesto}~\cite{10.1007/978-3-642-28108-2_19} has stood the test of time as a rallying call for the community, facilitating the establishment of a data science tradition within BPM research and ultimately anticipating the emergence of process mining offerings as important components in large ERP vendors' software suites.
In the realm of BPM and AI, the community has developed several conceptual visions.
Pre-dating widespread LLM adoption and new-generation AI agents, the manifesto on \emph{AI-augmented Business Process Management Systems} (ABPMS) focuses on the integration of AI capabilities into holistic process management and execution systems~\cite{dumas2023ai}, introducing the vision of framing in BPM while leaving the agentic perspective \emph{implicit}.
As a complementary viewpoint, the \emph{augmented process execution} proposal is intelligence/analytics-oriented and provides a four-level pyramid, ranging from basic \emph{descriptive} analytics via \emph{prediction} and \emph{prescription} to augmentation~\cite{DBLP:journals/sosym/ChapelaCampaD23}.
Paralleling the emergence of LLM-augmented process management software, the \emph{Large Process Models} vision sketches and discusses AI-based BPM in the age of LLMs, highlighting LLM potential to facilitate some tasks, while also emphasizing that achieving self-improving processes (essentially in the sense of an ABPMS) requires, even in human-in-the-loop scenarios, more than current-state LLMs~\cite{largeprocess}. 

Overall, the following four vision statements have anticipated at least some ideas that are largely accepted in practice:
\begin{enumerate}
    \item Intelligent business process execution requires the collaboration of autonomous software agents and humans; 
    \item These agents require symbolic \emph{frames};
    \item While LLM-based process modeling and mining capabilities play a role in mainstream BPM software offerings, LLMs alone are insufficient for truly intelligent BPM systems.
    \item More broadly, LLM-based technologies are driving the \emph{autonomous enterprise} era through novel agentic AI platforms, making it crucial to examine their technological implications for agentic BPM platforms as a central backbone.
\end{enumerate}
Continuing this tradition, we hope that our manifesto will both anticipate developments in BPM theory and practice and help researchers and practitioners embrace recent technological developments while planning for long-term sustainable impact on the discipline and the organizations that practice it.
In this context, we highlight the interdisciplinary nature of APM research as a bridge between BPM, autonomous agents, and machine learning. This space is being significantly reshaped by the recent trend of LLM-based agents, often portrayed in industry as a silver bullet, an innovation wave that the BPM community should actively engage with.

Parts of the APM vision can be practically realized by applying recent advances in research areas such as declarative process specification and conversational process modeling and mining.
Still, to fully realize APM systems, this manifesto should primarily be interpreted as a call to action for further research.
Notably, we intentionally leave open the realization and implementation details of key functionalities such as process-awareness and framing, an agent’s framed knowledge and goals representation, as well as the aforementioned agent capabilities for explainability, conversational actionability, and self-modification. We invite diverse concrete implementations of the presented vision and conceptual architecture.

Future work could address challenges along (but not limited to) the following lines:
\begin{description}[wide=0\parindent]
    \item[Establish formal abstractions for frames and goals.] To better support framed autonomy in BPM, we suggest to \emph{(i)} introduce first-class abstractions for mental models of processes, goals and normative frames; \emph{(ii)} develop and evaluate algorithms for synthesizing provably frame-compliant and performant operational specifications from frames, goals, and environmental information; \emph{(iii)} demonstrate the applicability of the abstractions and algorithms in the context of real-world business information systems.
    \item[Ensure explainability of \emph{execution} and \emph{management}.] Explainability for APM systems requires capabilities of technically sound explanations on process, agent, and MAS levels that are dynamically adjusted such that they are interpretable for the potentially heterogeneous group of target agents.
    The explanations should be actionable on both instance (execution) and process (management) levels as a means to maximize agentic autonomy.
    \item[Ensure actionability of conversational approaches.] Ensuring that agents behave faithfully and efficiently with respect to process goals requires research on how software agents can converse with human principals and other software agents in a process context, as well as with tools and traditional process-aware systems, how to assess agent behavior, and how to trade off autonomy and control.
    \item[Move self-adapting agents towards enterprise reality.] The deployment of concurrently evolving, process-aware, and self-adaptive agents in real-world enterprises requires new research on the governance of self-adaptation and systematic frameworks for the comprehensive assessment of the consequences of self-adaptation.
\end{description}
In this context, we advocate for research that \emph{(i)} integrates perspectives from both BPM and agents research; \emph{(ii)} advances strong formal foundations that give principle-based guarantees, \emph{(iii)} develops engineering results based on re-usable software artifacts and reproducible empirical evaluations grounded in systematic benchmarks; \emph{(iv)} ultimately studies the value provided to businesses and society at large.
Finally, we emphasize that a sustainable roadmap towards APM requires considering \emph{agent} notions that encompass not only LLM-based ``genAI agents'' but also include human agents that remain crucial for organizational and societal success.

\rmv{From an enterprise perspective, the need to develop enterprise-grade agents that deliver tangible business value represents an emerging and underexplored area of research. \rmv{AI-enabled and autonomous agents differ fundamentally from traditional automation systems, making it essential} The novel capabilities of AI-enabled and autonomous agents make it essential to identify the types of ``killer applications'' that could motivate enterprises to adopt such technologies within their core operations. Equally important is ensuring that these systems achieve the levels of reliability and profitability required in enterprise contexts. This topic warrants further investigation—not only from a technical standpoint but also from a business perspective, as well as for society at large.}

\subsubsection*{Acknowledgments}
\label{sec:acks}
The article was derived from working group notes as published in the report of Dagstuhl Seminar 25192, ``AUTOBIZ: Pushing the Boundaries of AI-Driven Process Execution and Adaptation''; these notes were further refined into several peer-reviewed workshop papers for the PMAI workshop at ECAI 2025~\cite{calvanese2025autonomy,Fettke2025,Montali25,Senderovich25}, which form the basis of Sections~\ref{sec:capabilities} and~\ref{sec:challenges} of this paper.

AC conducted this research while enrolled in the Italian National Doctorate in Artificial Intelligence run by Sapienza University of Rome (CUP B53C23003500006, Mission 4 NRRP Generic Research Scholarship, Component 1).
TK was partially supported by the Wallenberg AI, Autonomous Systems and Software Program (WASP) funded by the Knut and Alice Wallenberg Foundation.

\bibliographystyle{elsarticle-num} 
\bibliography{references,references-modif,tosem-refs}
\end{document}